\documentclass[10pt,twocolumn,letterpaper]{article}

\usepackage{my_style}
\usepackage{times}
\usepackage{epsfig}
\usepackage{graphicx}
\usepackage{amsmath}
\usepackage{amssymb}
\usepackage{verbatim}

\usepackage[pagebackref=true,breaklinks=true,letterpaper=true,colorlinks,bookmarks=false]{hyperref}

\Myfinalcopy 

\begin{document}

\title{Person Re-identification with Bias-controlled Adversarial Training}

\author{Sara Iodice and Krystian Mikolajczyk\\
Imperial College\\
London, UK\\
{\tt\small {\{s.iodice16,k.mikolajczyk\}}@imperial.ac.uk}
}

\maketitle

\begin{abstract}
   Inspired by the effectiveness of adversarial training in the area of Generative Adversarial Networks we present a new approach for learning feature representations in person re-identification. We investigate different types of bias that typically occur in re-ID scenarios, i.e., pose, body part and camera view, and propose a general approach to address them. We introduce an adversarial strategy for controlling bias, named Bias-controlled Adversarial framework (BCA), with two complementary branches to reduce or to enhance bias-related features. 
The results and comparison to the state of the art on different benchmarks show that our framework is an effective strategy for person re-identification. The performance improvements are in both full and partial views of persons. 
\end{abstract}

\section{Introduction}
Convolutional Neural Networks (CNNs) with recent innovations such as GANs~\cite{liu2018pose,wei2018person,deng2018image,zhong2017camera} and attention based models~\cite{li2018harmonious,xuattention,sidual} have fostered rapid development in human re-identification obtaining impressive results. 
However, these re-ID methods rely on CNN features which are susceptible to bias that the training data suffers from. Bias occurs when similarity between two data samples in re-ID problem is due to the same pose, body part or camera view, rather than to the ID-related cues.
For example,~\cite{geirhos2018imagenet} recently demonstrated that popular CNN architectures, i.e., ResNet-50 focus on local textures rather than global shapes for recognizing objects in ImageNet. It was also observed~\cite{song2018mask} that the background, i.e., camera view significantly affects feature representations in person re-ID when training on existing academic datasets. Other works~\cite{liu2018pose,su2017pose} deal with pose bias by generating multiple synthetic poses for training~\cite{liu2018pose} or by leveraging body part cues~\cite{su2017pose}. In contrast to all these methods that focus on one specific bias only, we propose a general approach that can address different types of bias without additional training data. 
\begin{figure}[h]
\includegraphics[height=6cm]{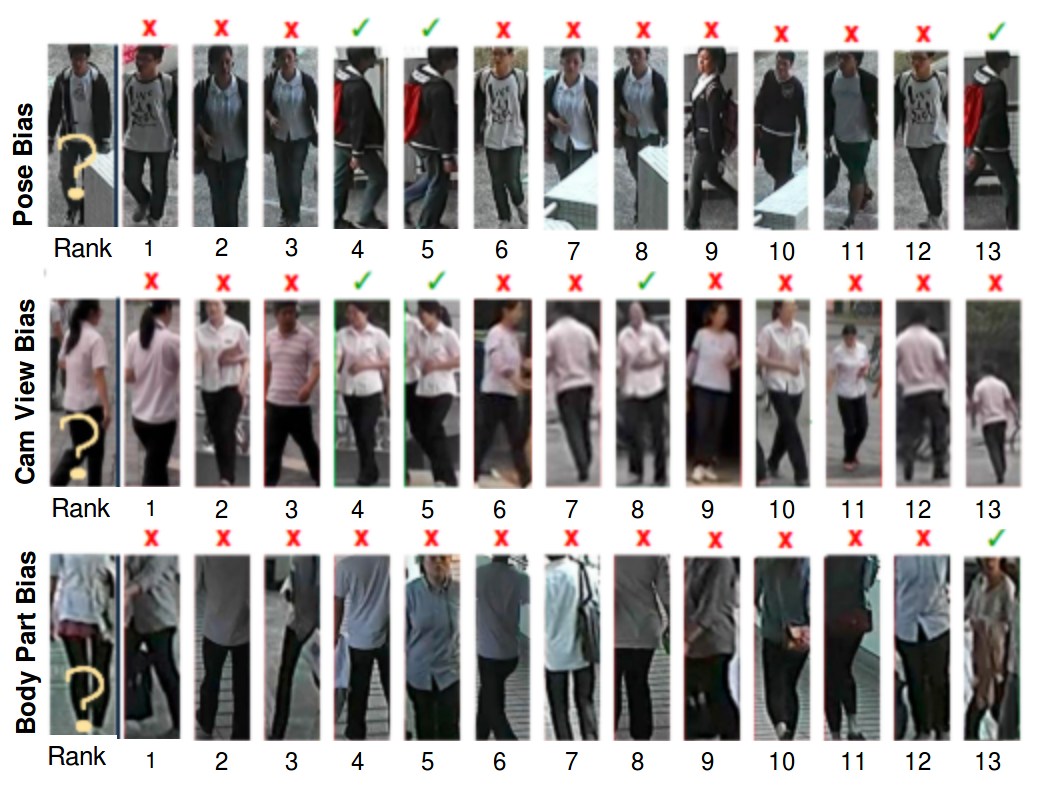}
   \caption{Examples of how different types of bias, i.e., pose, camera view, and body part affect the re-ID rankings.  We rank all gallery pictures by the distance to the query (from left to right). Frames containing different IDs are indicated with a red {x} above the frame. { Top row examples are from CUHK03, middle from Market-1501 and  bottom from CropCUHK03.}}
\label{fig:short}
\end{figure}
In figure \ref{fig:short} we show examples of how pose, camera view, and body part affect the re-id rankings. These  are top ranked returns given by a well performing TriNet~\cite{hermans2017defense} on   CUHK03~\cite{li2014deepreid}, Market-1501~\cite{zheng2015scalable} and CropCUHK03~\cite{iodice2018partial} with partial views. 
Frames with different IDs are marked by a red {x} above the frame. The top row shows  the correct ID  ranked $4$th after three different subjects in the same pose as the query; the second row illustrates camera view bias, where the correct ID is ranked next to similar subjects appearing in the same background as the query; finally,  the bottom row shows body part bias where the same body crop/part is ranked higher than the correct ID, which is ranked $13$th. 
Note that in person re-ID benchmarks the positive examples with the same bias are not included when calculating the performance score. 
This, however, can also be considered a certain type of bias as it does not exactly correspond to the practical scenario. Nonetheless, in re-ID settings, the bias shows with the top-ranked negative examples affecting re-ID performance as illustrated in figure~\ref{fig:short}.
Its influence can be reduced by large and diverse training data. However, in existing benchmarks, the small number of training examples often reflects specific settings from data collection. 
In this paper, we present an adversarial framework for controlling the bias in re-ID representation. 
Specifically, our model includes a feature mapping combined with an identity and a bias discriminators, that exploit examples with the same ID or the same bias. 
Our framework contains two specialized branches: one for {bias-reducing} in the image representations, e.g., pose, body part, background, thus focusing on ID features, e.g., clothes details; and one for {bias-enhancing} required for correct separation from ID features. 
The two branches generate descriptors containing complementary information to produce the final image representation. 
The contributions of this work are the following:
1) We introduce the idea of bias discriminator that helps to extract robust descriptors in re-ID;
specifically, we propose bias-controlled adversarial training with two network branches specialized in revealing bias-related and bias-unrelated structures in data, which generalizes existing adversarial solutions~\cite{liu2018exploring,ge2018fd} to other types of bias in human re-identification.  
2) This is the first work that provides quantitative analysis on how bias factors, such as the pose, camera view and body parts affect re-ID performance.  
3) We demonstrate experimentally that the two proposed branches {effectively reduce distracting features and enhance essential re-ID cues}. We show the importance of including both bias and ID features in the re-ID task leading to competitive results in several benchmarks with full and partial views. The two branches generate descriptors containing complementary information to produce the final image representation.

%

\section{Related work}
In this section, we review closely related methods and discuss the differences to our approach. 
\newline
\noindent\textbf{Background and Pose Bias.}
Despite many methods~\cite{li2017learning,zhao2017spindle,su2017pose,song2018mask} have proposed to leverage different regions/pixels to obtain descriptors robust to pose and background, the bias problem has not been explicitly analyzed and addressed. 
The only work investigating the influence of background is~\cite{tian2018eliminating} with the support of foreground masks. Instead, our framework relies on available annotations and low inference complexity compared to pixel based predictions. \newline
 \noindent\textbf{GANs based methods}~\cite{zheng2017unlabeled,su2017pose,liu2018pose,zhong2017camera,wei2018person,deng2018image} can potentially alleviate the bias problem by generating many synthetic samples with diversity, for example, in different poses~\cite{su2017pose,liu2018pose}, camera views~\cite{zhong2017camera}, and domains~\cite{wei2018person,deng2018image}. 
 However, none of these methods provides quantitative analyses of the bias problem in re-ID. \newline
 \noindent\textbf{Adversarial Training in Recognition.}
 Adversarial learning was previously adopted for face and person re-ID~\cite{liu2018exploring,ge2018fd} with similar goal of disentangling representations from biased features. 
  However, unlike our work,~\cite{ge2018fd} addressed the pose bias only with a specific solution relying on a pose landmark map as input for training. Furthermore, an unsupervised method~\cite{liu2018exploring} learns ID related and unrelated features for face identification, but its generalization to articulated pedestrian bodies is not straightforward. Our model relies on training data from standard datasets 
  only and leads to a better performance than~\cite{liu2018exploring}, as shown in the experimental results. 
\newline
\noindent\textbf{Adversarial Training in Domain Adaptation} from labelled source to unlabelled target has been proved to be particularly effective.  
According to the domain adaptation principles~\cite{ben2007analysis}, predictions should be made on features that can not discriminate between the source and the target domains. Thus~\cite{ganin2014unsupervised,ganin2016domain,tzeng2017adversarial} incorporate adversarial training to promote features that are discriminant for the main task in the source domain and non-discriminant with respect to the shift between the domains.
Likewise, we adopt adversarial training to extract features that are discriminant for the re-ID task, and non-discriminant with respect to the bias classes.
\newline
\noindent\textbf{Partial re-ID} is an emerging problem introduced in~\cite{zheng2015partial}, which focused on recognizing or matching identities of people from frames containing only parts of human bodies. 
A local-to-local matching based on small patches and global-to-local with a sliding window search for partial templates was used in~\cite{zheng2015partial}.  
Another approach~\cite{he2018deep} reconstructs missing channels in the query feature maps from full observations in the gallery maps. 
However,~\cite{zheng2015partial} was validated on small datasets, i.e., Partial Reid, i-LIDS, and Caviar, and may not scale well to real scenarios due to $On^4$ complexity during inference time, while~\cite{he2018deep} relies on the assumption that gallery images always contain full body of persons. 
Recently,~\cite{iodice2018partial} proposed a solution based on alignment and hallucination as well as a synthetic data generated from CUHK03 with partial views in both probe and gallery images.
Similarly, in order to analyze partial view bias, we evaluate our approach on CropCUHK03 dataset~\cite{iodice2018partial} and compare with their proposed method.

\section{Adversarial training}
In this section, first we shortly summarise the main concepts our work is based on, and then present the proposed bias-controlled adversarial framework.

\noindent\textbf{\subsection{Preliminaries}}
\noindent\textbf{Adversarial Training Framework.} The goal of the {generative adversarial framework}~\cite{goodfellow2014generative} is to train a differentiable function $G_W$ parametrized by $W$, usually represented by deep neural networks, to generate fake samples 
as close as possible to the real data. $G_{W}$ is trained along with a discriminator $D_{\Phi}$, which estimates the likelihood that an input sample comes from the real distribution. $D_{\Phi}$ is adversarial to $G_W$ in the sense that by recognizing fake from real samples, $D_{\Phi}$ drives $G_W$ to generate more realistic examples. 

\begin{figure*}[h]
\centering
\includegraphics[width=1.05\textwidth,height=170 pt]{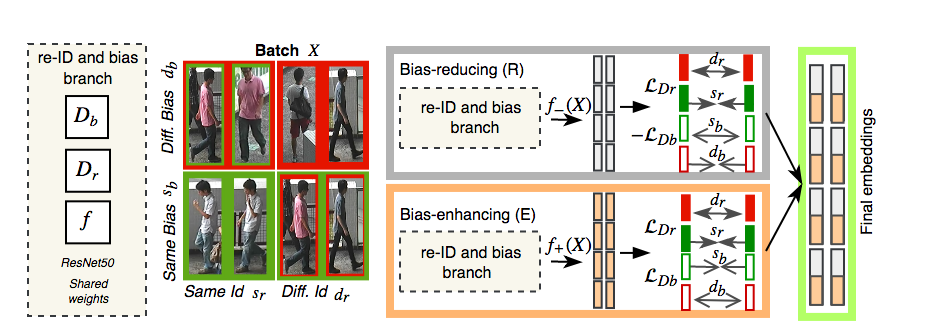}
\caption{Bias-controlled Adversarial framework (BCA) includes two independent branches, i.e., bias-reducing (R) and bias-enhancing (E) sharing the same architecture but different weights. Each re-ID and bias branch includes a feature mapping $f$ and discriminators $\{D_r,D_b\}$ sharing the weights of ResNet50 backbone. {In each branch, the feature mapping $f$ is driven by $D_r$ and $D_b$ to reduce or enhance bias by changing the sign of $\mathcal{L}_{Db}$. $D_r$ and $D_b$ are trained by triplets, where $s_r,d_r$ are the positive/negatives pairs selected by considering their ID, and $s_b,d_b$ by considering their bias type.   
During inference time, the final descriptor is obtained by concatenating the two embeddings given by the mappings $f$ of  the two independent branches.}}
\label{fig:Adversarial_training}
\end{figure*}
This is known as {$\min\max$ game} with value function $L(G,D)$:

\begin{equation} \begin{split}
    \min_{W} \max_{\Phi}L(G,D)&= \mathbb{E}_{x \sim p_{r}(x)}[{\log D_{\Phi}({x})}] \\ &+ \mathbb{E}_{z \sim p_{z}(z)} [\log (1-D_{\Phi}(G_{W}(z))]
    \end{split}
    \label{eq:1}
\end{equation}
${x}$ and $G_W(\cdot)$ are real and fake samples from input noise $z$, respectively. It results in the following adversarial loss:
\begin{equation} \begin{split} 
    \mathcal{L}_{adv}= -\mathbb{E}_{x \sim p_{r}(x)}[{\log D({x})}]  - \mathbb{E}_{x \sim g(x)} [\log (1-D(x))]
      \end{split}
    \label{eq:2}
\end{equation}
Many variants of GANs have been proposed with modifications to the original framework  making it a powerful and flexible tool, e.g., ~\cite{durugkar2016generative} introduced multiple discriminators:
\begin{equation}
{\begin{split}  
   \mathcal{L}_{adv}&= \\&\sum_{i=1}^n -\mathbb{E}_{x \sim p_{r}(x)}[{\log D_i({x})}]  - \mathbb{E}_{x \sim g(x)} [\log (1-D_i(x))]
 \end{split} }
 \label{eq:3}
\end{equation}
In domain adaptation problem,~\cite{tzeng2017adversarial} replaces the generator with the mapping $f$ for making target samples $X_t$ indistinguishable from those in the source distribution $X_s$:
\begin{equation} 
\begin{split} 
   \mathcal{L}_{adv}(f)=& -\mathbb{E}_{x \sim X_s}[{\log D({f(x)})}] \\& - \mathbb{E}_{x \sim X_t} [\log (1-D(f(x))]
\label{eq:3domain}
\end{split} 
\end{equation}
 A new objective, as an  alternative to the binary cross entropy, was proposed in~\cite{salimans2016improved}. It is based on similarity between distributions of the generated and real data:
\begin{equation}
\begin{split} 
   \mathcal{L}_{adv}=  ||\mathbb{E}_{x \sim p_{r}(x)}f(x)  - \mathbb{E}_{z \sim g(z)} f(G(z))||_2^2
\label{eq:gan_real}
\end{split} 
\end{equation}


Inspired by these methods, we 1) use multiple discriminators~\cite{durugkar2016generative} to recognize samples with the same identity or bias; 2) instead of employing a generator, we train feature mappings as in~\cite{tzeng2017adversarial} with controlled bias; 3) we train the feature mappings as~\cite{salimans2016improved} to directly match the corresponding feature vector of the correct sample, instead of using the typical cross entropy loss.

\noindent\textbf{Triplet Loss.}~\cite{weinberger2009distance,ding2015deep,oh2016deep} Training samples are selected  and arranged in triplets $x_a,x_p,x_n$, where
$x_a,x_p$ is the pair with the same identity therefore their distance is minimized and $x_a,x_n$ is the pair with different identities therefore their distance is maximized.
\newline
\textbf{Batch Hard}~\cite{hermans2017defense} has lead to a superior performance compared to the triplet loss~\cite{ding2015deep,oh2016deep}. Positives and negatives are paired in a batch specifically designed to include $C$ random classes, with randomly sampled $K$ examples of each class, thus resulting in a batch of $CK$ images. For each anchor in the batch the hardest positive
and the hardest negative sample is selected when forming triplets.
This leads to the following formulation of loss with mini-batch $X$ and margin $m$:
%
%
\begin{equation}
{\begin{split} \mathcal {L}(X)= \sum_{x_a\in X} [ m+\overbrace{\max_{x_p}||x_a-x_p||^2 }^{hard\,positive} - \overbrace{\min_{{x_n }} ||x_a-x_n||^2}^{hard\,negative}]_+
   \label{eq_6}
    \end{split}}
\end{equation}
\subsection{Proposed model}
We introduce our Bias-controlled Adversarial framework (BCA) with two main branches for bias-reducing (R) and bias-enhancing (E). The former suppresses features corresponding to the body part/pose/camera bias and selects ID features that are robust against their variations. The latter enhances features relevant for pose, camera or body part recognition.
A overview of the training framework is presented in figure \ref{fig:Adversarial_training}.
Each branch includes two discriminators $D_r$ and $D_b$ and the feature mapping $f$. 
While the re-ID discriminator $D_r$ attempts the re-ID task, the bias discriminator $D_b$ drives the feature mapping $f$ to reduce/enhance the bias features by minimizing the proposed adversarial triplet loss.
Both branches are trained with adversarial loss for re-ID and for bias discriminator as presented below.\newline
{\noindent\textbf{Re-ID Adversarial loss.}} We define the re-identification objective $\mathcal{L}_{Dr}$ as an adversarial triplet loss, which is a reinterpretation of batch hard triplet loss in equation \ref{eq_6} including GANs equations \ref{eq:3domain} and \ref{eq:gan_real}:{ \begin{equation}\small{\begin{split} \mathcal{L}_{Dr}(X)&= \\& \sum_{a_r\in X} [ m+\overbrace{\max_{p_r}||\mathbb{E}_{a_r \sim p(a_r)}f(a_r)  - \mathbb{E}_{p_r \sim p(p_r)} f(p_r)||^2}^{hard\,positive}\\& - \overbrace{\min_{{n_r }} || \mathbb{E}_{a_r \sim p(a_r)}f(a_r)  - \mathbb{E}_{n_r \sim p(n_r)} f(n_r)}^{hard\,negative}||^2]_+
   \label{eq_7_ReIDAd}
    \end{split}}
\end{equation}}
where parameters of the feature mapping $f$ are optimized so that the distribution of positive ID class samples $p(p_r)$ matches the distribution of anchor ID class samples $p(a_r)$, and also differs from the distribution of negative ID samples $p(n_r)$. \newline
\noindent\textbf{Bias Adversarial loss.} 
Similarly to equation \ref{eq_7_ReIDAd}, bias adversarial loss is defined as:
{
\begin{equation}\small{\begin{split} \mathcal {L}_{Db}(X)&=\\& \sum_{a_b\in X} [ m+\overbrace{\min_{p_b}||\mathbb{E}_{a_b \sim p(a_b)}f(a_b)  - \mathbb{E}_{p_b \sim p(p_b)} f(p_b)||^2}^{easy\,positive}\\& - \overbrace{\max_{{n_b }} || \mathbb{E}_{a_b \sim p(a_b)}f(a_b)  - \mathbb{E}_{n_b \sim p(n_b)} f(n_b)}^{easy\,negative}||^2]_+
   \label{eq_7}
    \end{split}}
\end{equation}
}
In contrast to  $\mathcal{L}_{D_r}$, training  triplets $\{p_b,a_b,n_b\}$ for $\mathcal{L}_{D_b}$ are selected by considering the bias types and the easiest positive/negative samples as more meaningful to the bias term. In other words, for the re-ID task, it is more beneficial to include hard cases in the batch, e.g, same persons with a different pose, or similar pose of different persons. Instead, for the bias term, the easiest cases are those where the bias is more evident and therefore should be used for learning bias related characteristics.
However, there are no specific constraints on the ID when pairing samples for the bias. 
{\noindent\textbf{Adversarial Triplet Loss.}}
Our objective evaluates the loss for a mini-batch of samples $X$ arranged in triplets with the batch hard strategy. It involves the bias discriminator $D_b$, the ID discriminator $D_r$ and the mapping functions $f_{\circ}$: 
\begin{equation}
\mathcal{L}_{adv}(X)=\lambda_{Dr} \cdot \mathcal{L}_{D_r}(f_{\circ}(X))\circ \lambda_{Db} \cdot \mathcal{L}_{D_b} (f_{\circ}(X)) 
\label{eq:atl}
\end{equation}
$\mathcal{L}_{adv}(X)$ incorporates re-ID adversarial loss $\mathcal{L}_{Dr}$ and bias adversarial loss $\mathcal{L}_{Db}$ for training $D_r$ and $D_b$.  
Note that the loss corresponds to both the bias-reducing and bias-enhancing branch, depending on the sign ($\circ$) between the two components.
Maximizing  $\mathcal{L}_{Db}$ loss (-) reduces features related to bias in mapping images to features by $f_{-}$, while minimizing $\mathcal{L}_{Db}$ (+)  leads to bias enhancing in $f_{+}$.  
Parameters $\lambda_{Dr}$ and $\lambda_{Db}$  control the contributions from these two terms in each branch.
\newline
{\noindent\textbf{Bias-reducing branch }} is obtained by setting sign ($\circ$) to ($-$) in Equation \ref{eq:atl},
thus $\mathcal{L}_{Db}$ is maximized, i.e., the distance between samples from the same bias class $a_b,p_b$, regardless their ID is maximized, while distance between different bias class samples $a_b,n_b$ is minimized.
This forces $f_{-}$ to reduce feature components encoding the bias and focus on unbiased patterns.
\newline
\noindent\textbf{Bias-enhancing branch.}
$\mathcal{L}_{Db}$ is minimized by setting sign ($\circ$) to ($+$) in Equation \ref{eq:atl}, i.e., the distance between $a_b$ and $p_b$ is minimized, and between $a_b$ and $n_b$ is maximized. The network rewards biased samples, therefore emphasizes feature components encoding the bias. 
As we show in the experiments, it provides a complementary mapping branch $f_{+}$ that improves the representation for re-ID task. 
\newline
Re-ID loss $\mathcal{L}_{Dr}$ is used in training in the same way in both scenarios to reduce the distance between positive ID pairs and increase for the negative ones.  In contrast, the bias loss $\mathcal{L}_{Db}$ differs for the two cases i.e., maximizing or minimizing the distance between positive bias pair for reducing and enhancing branches, respectively. Finally, during inference time the two embeddings - coming from the mappings $f_{-}$ and $f_{+}$ from the two independent branches - are concatenated to produce the final image representation.

\section{Experimental results} \label{experimental_results}
In this section, we first describe datasets used for our evaluation and provide implementation details. We then analyze the effectiveness of our BCA framework for different types of bias. Finally, we report the performance on standard benchmarks and compare our approach against other methods on full and partial views. 
\subsection{Datasets and bias annotation}
\noindent\textbf{CUHK03}~\cite{li2014deepreid} includes $14,096$ images of $1,467$ pedestrians from 6 surveillance cameras. We adopt the new evaluation protocol from~\cite{zhong2017re}. In this dataset we experiment with the pose bias considering two class labels, i.e., frontal and profile views, which also correspond to the two disjoint camera views.
\newline
\noindent\textbf{MARKET-1501 }~\cite{zheng2015scalable} contains $32,668$ annotated bounding boxes of $1,501$ identities. The images are acquired from six surveillance cameras including $5$ high-resolution cameras and $1$ low-resolution camera. In Market-1501 we investigate both the camera view and pose bias using the available camera annotation and the proposed three class pose labels, i.e., frontal, side and oblique.
\newline
\noindent\textbf{DUKE}~\cite{ristani2016MTMC} provides $36,411$
bounding boxes of $1,501$ identities. As in Market-1501, we analyze both the camera view and the pose bias from the existing camera annotation and our three pose classes.
\newline
\noindent\textbf{CropCUHK03}~\cite{iodice2018partial} is a  dataset derived from CUHK03~\cite{li2014deepreid} by cropping partial views but maintaining the same number of individuals and frames. 
We use CropCUHK03 settings $_{s=0.5,o_{min}=.25}$ where $s$ is the fraction of area cropped from full labeled frame and $o_{min}$ is the minimum overlap between two views. The pose labels are frontal and side. In addition, we annotate each partial view with three body part labels, i.e., upper, central, bottom parts.  \newline
Note that in all the datasets class labels are uniformly distributed.
Our pose annotation for Market-1501 and Duke as well as body part annotation for CropCUHK03 are available online\footnote{https://github.com/iodicesara/person-re-identification-with-bias-controlled-adversarial-training}.
\newline
\subsection{Implementation details}
\noindent\textbf{Baseline.} We use an implementation of TriNet~\cite{hermans2017defense} as a baseline for comparison in our experiment.
We set a batch size of $P\cdot K=64$ samples with $P=16$ randomly extracted persons IDs and $K=4$ instances for each person. We train the network with Adam optimizer, with a learning rate of $0.0003$ and linear decay to zero over $60$ epochs. As common practice during the training, we perform data augmentation by using random horizontal flips and re-scaling input images to $384\times128$ pixels. 
Further implementation details can be found in~\cite{hermans2017defense}.\newline
\noindent\textbf{Bias-reducing/enhancing branches.}  ResNet50 is used as a backbone of  the discriminators $\{D_b,D_r\}$  and the mapping $f$ in each  of the two branches. In our implementation, these three components share the same weights within each of the ResNet50 branch. 
Each branch is trained independently and with the same hyper-parameters, optimizer and augmentation strategy as the baseline. {In bias-reducing, we set $\lambda_{D_b}=-0.01$ for all types of bias. In bias-enhancing, we set $\lambda_{D_b}=0.05$ for pose bias and $\lambda_{D_b}=0.01$ for camera and body part bias. As a good practice, we kept consistent across different datasets the values of $\lambda_{D_b}$.
During inference, mappings $f_{-}$ and $f_{+}$ output a concatenated descriptor of $2 \times2048$ components from the-second-to-last layers of ResNet50.} Note that no bias annotation is required during inference time.
\newline
\begin{table}[h]
\scriptsize
\begin{tabular}
{@{\hskip4pt}c@{\hskip4pt}c@{\hskip5pt}c@{\hskip5pt}c@{\hskip5pt}c@{\hskip5pt}c@{\hskip5pt}c@{\hskip5pt}c@{\hskip5pt}c@{\hskip5pt}c@{\hskip5pt}c@{\hskip5pt}c@{\hskip5pt}c@{\hskip5pt}c@{\hskip5pt}c@{\hskip5pt}c@{\hskip5pt}c@{\hskip5pt}c@{\hskip5pt}c@{\hskip5pt}c} 
\multicolumn{1}{c}{\textbf{}} & \multicolumn{3}{c}{{Pose}} & \multicolumn{3}{c}{{Camera}} & \multicolumn{3}{c}{{BodyPart}} \\ 
\multicolumn{1}{c}{\textbf{}} & \multicolumn{3}{c}{{CUHK03}} & \multicolumn{3}{c}{{Market-1501}} & \multicolumn{3}{c}{{CropCuhk03}} \\\hline
 & \multicolumn{2}{c}{re-ID} & \multicolumn{1}{c|}{Bias} & \multicolumn{2}{c}{re-ID} & \multicolumn{1}{c|}{Bias} & \multicolumn{2}{c}{re-ID} & Bias \\
 & {Rank1} & mAP & \multicolumn{1}{c|}{acc} & Rank1 & mAP & \multicolumn{1}{c|}{{acc}} & Rank1 & mAP & {acc} \\
{Bas} & {60.7} & {72} & \multicolumn{1}{c|}{{.78}} & {87} & {72.9} & \multicolumn{1}{c|}{{.44}} & {39.6} & {52.6} & {.56} \\
{NoBias} & {\textbf{63.5}} & {74} & \multicolumn{1}{c|}{-} & {\textbf{88.6}} & {75.3} & \multicolumn{1}{c|}{-} & {\textbf{60.5}} & {\textbf{69.8}} & \multicolumn{1}{c}{-} \\ \hline
\multicolumn{1}{c}{$-\lambda_{Db}$} & \multicolumn{9}{c}{{Bias-reducing (R) $(\lambda_{Dr}=1)$}} \\ \hline
$.1$ & {26.5} & {39.6} & \multicolumn{1}{c|}{\textbf{.50}} & {79.2} & {64.4} & \multicolumn{1}{c|}{{.22}} & {15.6} & {26} & {.53} \\
$.05$ & {56.2} & {68.3} & \multicolumn{1}{c|}{\textbf{{.51}}} & {83.6} & {68.7} & \multicolumn{1}{c|}{{.22}} & {26.1} & {38.4} & {.53} \\
$.01$ & \textbf{{62.5}} & {73.3} & \multicolumn{1}{c|}{{.62}} & \textbf{{87.2}} & {72.9} & \multicolumn{1}{c|}{{.27}} & \textbf{{40.6}} & \textbf{53.6} & {.54} \\
$.005$ & {61.7} & {72.9} & \multicolumn{1}{c|}{{.71}} & {86.8} & {73.2} & \multicolumn{1}{c|}{.31} & {38.8} & {51.8} & {.54} \\
\hline
\multicolumn{1}{c}{$\lambda_{Db}$} &\multicolumn{9}{c}{{Bias-enhancing (E) $(\lambda_{Dr}=0)$ }}
\\ \hline
$1$ & {.2} & {1.3} & \multicolumn{1}{c|}{{1}} & {.0} & {.1} & \multicolumn{1}{c|}{{.82}} & {.2} & {1.1} & {.66} \\
\hline
\multicolumn{1}{c}{$\lambda_{Db}$}
&\multicolumn{9}{c}{{Bias-enhancing (E) $(\lambda_{Dr}=1)$ }} \\ \hline
$.005$ & {60.7} & {71.9} & \multicolumn{1}{c|}{{.77}} & {87.8} & {72.9} & \multicolumn{1}{c|}{{.44}} & \textbf{40.3} & \textbf{53.2} & {.56} \\
$.01$ & {61.2} & {72.4} & \multicolumn{1}{c|}{{.74}} & \textbf{88.5} & {73.8} & \multicolumn{1}{c|}{.43} & {39.9} & {52.7} & {.56} \\
$.05$ & \textbf{62.4} & {73.2} & \multicolumn{1}{c|}{{.69}} & {88.2} & {74.5} & \multicolumn{1}{c|}{{.43}} & {36.1} & {49} & {.55} \\
$.1$ & {61.4} & {72.4} & \multicolumn{1}{c|}{{.62}} & {87.4} & {72.9} & \multicolumn{1}{c|}{{.32}} & {21.9} & {33.5} & {.53} \\ \hline
{R+E} & {\textbf{65.1}} & {75.3} & \multicolumn{1}{c|}{-} & {\textbf{89.9}} & {77.1} & \multicolumn{1}{c|}{-} & {\textbf{42.1}} & {\textbf{54.7}} & \multicolumn{1}{c}{-}
 \\ 
\hline

\end{tabular}
\caption{Analysis of re-ID  and bias classification task under different values of $\lambda_{Db}$ parameter for either reducing (R) or enhancing (E) bias features on CUHK03 (labeled split and new protocol), Market-1501 (single query), and CropCUHK03. Re-ID performance is measured in terms of Rank1 and mAP (mean average precision); Bias classification task is evaluated in terms of accuracy (acc).}
\label{tab:analysisbias}
\end{table}
\subsection{Evaluation of individual components}
\noindent We evaluate the effectiveness of our BCA framework for different bias, i.e., pose, camera, and body part.
We first demonstrate that {bias-reducing} and {bias-enhancing} branches effectively suppress and boost the bias features by evaluating the generated re-ID descriptors for bias classification tasks.
Furthermore, we show the importance of bias related features and how they affect the  representations that are essential for the re-ID task.  
Finally, we statistically analyze the occurrence of bias during evaluation and prove that our method is able to control it. 
The bias classification experiment  shows the extent to which the performance of a strong TriNet baseline~\cite{hermans2017defense} is affected by the pose, camera and body parts. 
We extract features with pretrained mappings $f_{\circ}$ from bias-reducing (R) and bias-enhancing (E) branches on three different datasets. CUHK03 for front/side pose recognition; Market-1501 for six cameras; CropCUHK03 for top, central and bottom body parts.
We use the extracted features as inputs to train a simple classification model based on a PReLU activation layer and a fully connected layer with a number of outputs corresponding to the bias class labels, i.e., $2$ for pose, $6$ for camera, and $3$ for body part. Note that while training the bias classifier the feature extracting backbone is fixed.
Table \ref{tab:analysisbias} shows the results which we discuss in detail below.  
\newline
\noindent\textbf{Baseline.} In order to measure the top performance the system can reach  we evaluate the baseline disregarding biased pose, camera or body part  samples. Specifically, during inference for each query we remove  the samples  from the gallery list that have the same bias label as the query and calculate the scores. 
\begin{figure*}[h]
\centering
{\includegraphics[height=8.5cm,width=17cm]{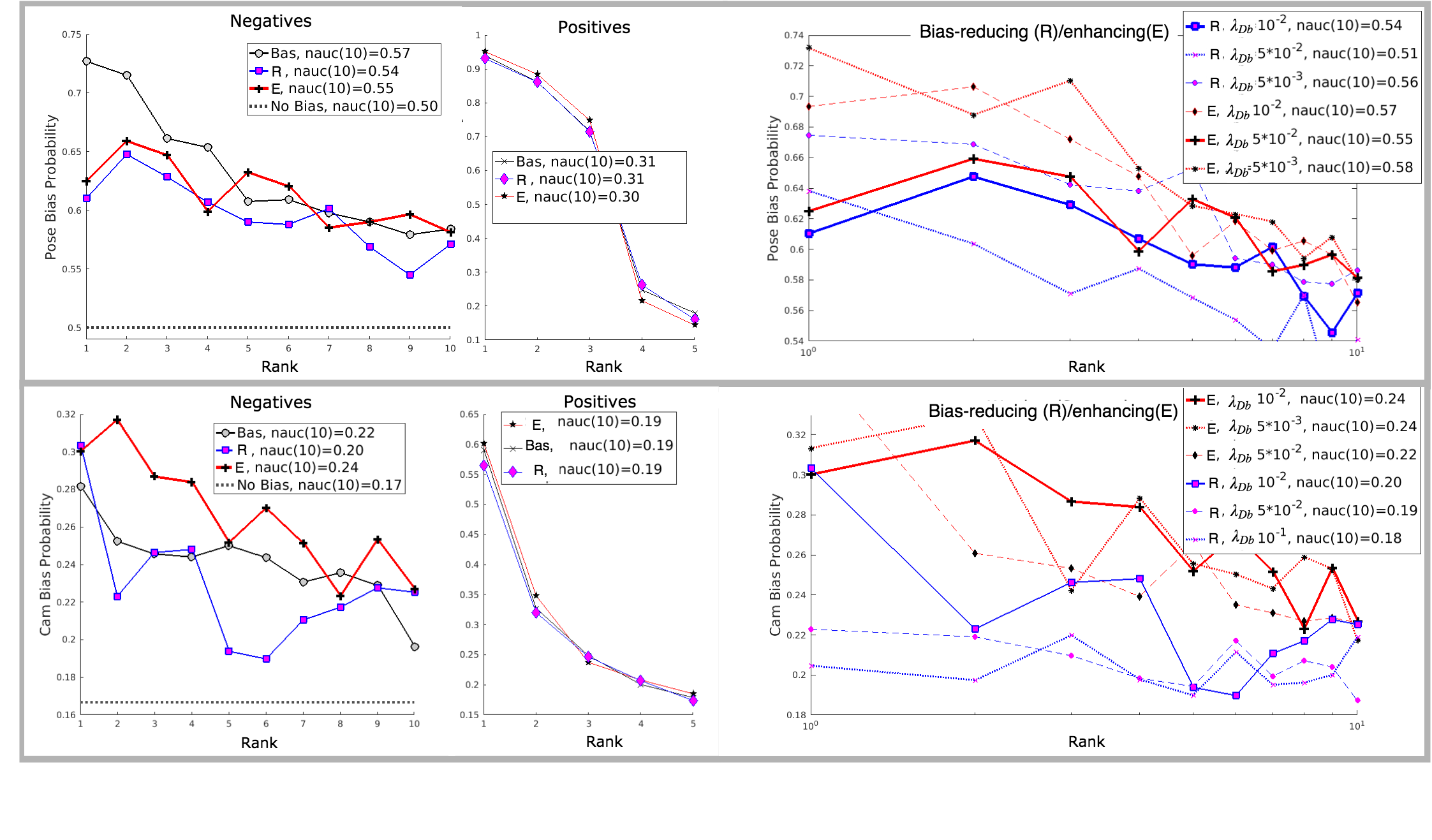}}
   \caption{Statistical analysis of bias under different configurations i.e. baseline (Bas), bias-reducing (R), bias-enhancing (E) on datasets CUHK03 (top) and Market-1501 (bottom) for pose and camera bias. Graphs illustrate the following statistics: left) the probability that a negative sample with the same bias appears at a certain rank position; middle) the probability that a positive sample with the same bias appears at a certain rank position; right) analysis of bias in negatives under different values of parameter $\lambda_{Db}$.}
\label{fig:statistic}
\end{figure*}
Note that, re-ID protocols already discard positives with the same camera view to make the task challenging and in this experiment we additionally exclude the negatives with the same bias. The results show that the re-ID performance  is significantly affected by the bias as we observe Rank1 improvement in each of the benchmarks by $2.8\%$ on CUHK03, by $1.6\%$ on Market-1501, and  by $20.9\%$ on CropCUHK03. 
The results also show that re-ID features from the baseline include information on specific bias as the bias classification accuracy is $0.78$, $0.44$ and $0.56$ for pose, camera and body parts, respectively. 
\newline
\noindent\textbf{Bias control.} Next we demonstrate the ability of the proposed framework to control the bias.  We train the two branches for different balance between the re-ID and the bias loss in equation~\ref{eq:atl}, i.e., we vary the values of $\lambda_{Db}$ for pose, camera and body parts while keeping re-ID loss fixed $\lambda_{Dr}=1$.
For example in table \ref{tab:analysisbias}, by setting  $\lambda_{Db}$ to  negative weight of $-0.01$ i.e.,  the bias classification accuracy is  reduced from $0.78$ to $0.62$ for the pose bias in CUHK03  and from $0.44$ to $0.27$ for the camera bias in Market-1501, with re-ID performance slightly increased. Further increasing negative $\lambda_{Db}$  deteriorates the re-ID performance. 
%
It is also evident that body part bias is more challenging to control than other types, however we are still able to reduce the bias classification accuracy in CropCUHK03 from $0.56$ to $0.54$ while increasing re-ID in both Rank1 and mAP metrics. 
{ Note that, re-ID performance is very sensitive to the value of $\lambda_{Db}$ and its behaviour is somewhat opposite in the two branches.
%
Moreover, when re-ID loss $\lambda_{Dr}$ is deactivated and $\lambda_{Db}=1$, features from Bias-enhancing branch well recognizes all the bias types. The classification accuracy is $1$ for pose, $0.82$ for six cameras, and $0.66$ for top, central and bottom body parts.  
These suggest that the bias features in re-ID descriptors and features learned from the bias labels are complementary which motivates us to include two different branches for controlling the bias in person representation.
The reducing one for disentangling the bias from re-ID features in order to obtain robust representation. The enhancing one for boosting features from the available bias annotation. Note that the importance of including bias related features was previously highlighted by other works which incorporate shape features from background masks~\cite{su2017pose} and body parts~\cite{song2018mask} into re-ID representations as related to essential cues for re-ID.
}
Our results further supports their observations.

\noindent\textbf{Effect of bias on re-ID.} 
An important observation under {bias-reducing} mode in table \ref{tab:analysisbias} is that by increasing the contribution from ${\lambda_{Db}=0.005}$ to ${\lambda_{Db}=0.1}$, Rank1 dramatically drops from $60.7\%$ to $26.5\%$ in CUHK03, from $87\%$ to $79.2\%$ in Market-1501 and from $39.6\%$ to $15.6\%$ with respect to the baseline. This suggests that pose, camera and body part  features cannot  be completely suppressed without affecting re-ID features. 
We found that effective bias reduction  is with $\lambda_{Db}\leq 0.01$ resulting in an improvement from $60.7\%$ to $62.5\%$ in CUHK03 and $39.6\%$ to $40.6\%$ in CropCUHK03.
\newline
Another observation is that by emphasizing the opposite configuration, which enhances the bias features related to the annotation, Rank1 increases by nearly $2\%$ in CUHK03 ($\lambda_{Db}=0.05$) and by $1.5\%$ in Market-1501 ($\lambda_{Db}=0.01$). Finally, concatenating the embeddings derived from both branches results in a considerable boost in Rank1 in all three benchmarks, i.e., from $60.7\%$ to $65.1\%$ in CUHK03, from $87\%$ to $89.9\%$ in Market-1501 and from $39.6\%$ to $42.1\%$ (case: $R+E$). This further suggests that {bias-reducing} and {bias-enhancing} branches extract complementary features and form a better descriptor for re-ID when used jointly.\newline
\noindent\textbf{Statistical analysis of bias in retrieval.} 
Ranked gallery samples returned by the re-ID system include both positive and negative examples. Re-ID evaluation protocols discard only the positives from the same camera, however, the negatives are also biased and affect the re-ID scores. 
We analyze the influence of pose and camera bias in the ranked lists. 
In figure \ref{fig:statistic} we report statistics under different configurations, i.e., baseline (Bas), bias-reducing (R), bias-enhancing (E) on datasets CUHK03 and Market-1501 for pose (top) and camera bias (bottom). The left and middle figures show the probability that a negative and positive sample, respectively, with the same bias appears at a certain rank position. We also report $nauc_{10}$ aggregating probability values over the first $10$ rank positions.
The bias is clearly visible for the baseline (black curve) significantly higher than the ideal case of no bias (grey dotted curve).  Bias-reducing branch (R) leads to a drop in $nauc(10)$ by $2\%$ from $0.57$ to $0.55$ for pose and from $0.57$ to $0.55$ for camera bias, which further shows its effectiveness.
In contrast, the bias-enhancing branch increases the probability of biased samples occurring higher in the rank, e.g., increasing values of $nauc(10)$.\newline
Similarly, we analyze the behaviour of bias in positives. Although they are not included in the evaluation protocol, we perform this experiment to confirm that the proposed approach does not penalize them, which would have a negative effect in a practical scenario where all retrieved positive samples matter. { This is clearly visible in figures \ref{fig:statistic} (middle) where the probability curves and $nauc(10)$ values remains nearly the same for baseline as well as for R and E branches.}\newline
{Finally, in figure \ref{fig:statistic} (right) we show that  the influence of  bias on the rankings can be controlled by varying the parameters $\lambda_{Db}$ in the bias-reducing and enhancing branches. For example, increasing $\lambda_{Db}$ from $5 \cdot 10^{-3}$ to $5 \cdot 10^{-2}$, leads the bias-reducing branch (blue curves) to decrease the $nauc(10)$ by $0.5$ in same pose probability (top) and by $0.2$ in same camera  probability (bottom). On the other hand, decreasing $\lambda_{Db}$ from $5 \cdot 10^{-2}$ to $5 \cdot 10^{-3}$ (red curves) in bias-enhancing boosts the probability to retrieve a biased sample from $0.55$ to $0.58$ for the pose and from $0.22$ to $0.24$ for the camera.}
\subsection{Comparative evaluation}
We compare our approach against the state of art methods, in particular GAN, attention, background bias, and other adversarial learning techniques in tables \ref{tab:compMark}, \ref{tab:compCUHK03} and \ref{tab:compcrop}  for Market-1501 and Duke, CUHK03 as well as CropCUHK03, respectively.  Our approach is referred to as R+E (bias reducing+enhancing) followed by bias type. In order to make it more complete we also indicate for each method if external data (D+), random erasing augmentation strategy (A+) and re-ranking (RR+) are adopted, as this practices positively affect the re-ID performance.
\begin{table}[h]
\footnotesize
\begin{tabular}
{lcl|clclcl}
 & \multicolumn{2}{c}{MARKET-1501} & \multicolumn{2}{c}{DukeMTCM} \\
Method & Rank1 & mAP & Rank 1 & mAP \\ \hline
LOMO+XQDA~\cite{liao2015person} & - & - & 30.8 & 17 \\
MSCAN~\cite{li2017learning} & 80.3 & 57.5 & - & - \\
Spindle~\cite{zhao2017spindle} & 76.9 & - & - & - \\
SVDNet~\cite{sun2017svdnet} & 82.3 & 62.1 & 76.7& 56.8 \\
PCB~\cite{sun2017beyond} & \textbf{93.8} & \textbf{81.6} & \textbf{83.3} & \textbf{69.2} \\ 
\hline
\textit{GAN} &  &  &  &  \\
DCGAN~\cite{zheng2017unlabeled} & 83.9 & 66.1 & 67.7 & 47.1 \\
Pose-transfer~\cite{liu2018pose} & 87.6 & 68.9 & 78.5 & 56.9 \\
CamStyle~\cite{zhong2017camera} & 88.1 & 68.7 & 75.3 & 53.5 \\
\textbf{CamStyle(RR+)}~\cite{zhong2017camera} &\textbf{89.5} & 71.5 & \textbf{78.3} & \textbf{57.6} \\
Pose-norm~\cite{qian2018pose} & 89.4 & \textbf{72.6} & 73.5 & 53.2 \\ 
\hline
\textit{Attention} &  &  &  &  \\
ACCN (D+)~\cite{xuattention} & 85.9 & 66.9 & 76.8 & 59.2 \\
ACCN (D+RR+)~\cite{xuattention} & 87.7 & \textbf{83.0} & - & - \\
HA-CNN~\cite{li2018harmonious} & 91.2 & 75.7 & 80.5 & 63.8 \\
DuATM (A+)~\cite{sidual} & 91.4 & 76.6 & 81.8 & 64.6 \\
\textbf{Mancs~\cite{wang2018mancs}}& \textbf{93.1} & 82.3 & \textbf{84.9} & \textbf{71.8} \\ \hline
\textit{Background bias} &  &  &  &  \\
E. Backgr-Bias~\cite{tian2018eliminating} & 80.5 & - & - & - \\ 
E. Backgr-Bias(D+)~\cite{tian2018eliminating} & 81.2 & - & - & - \\
MaskG\cite{song2018mask} & 83.8 & 74.3 & - & - \\
\hline
\textit{Adversarial learning} &  &  &  &  \\
DR-GAN (siam)~\cite{liu2018exploring} & 86.7 & 73.2 & 76.9 & 60.2 \\
\textbf{FD-GAN}~\cite{ge2018fd} & \textbf{90.5} & \textbf{77.7} & \textbf{80.0} & \textbf{64.5}\\ \hline
\textit{Our method} &  &  &  &  \\
baseline  & 87.0 & 72.9 & 78.4 & 63.6 \\
R+E pose & 89.9 & 77.3 & 80.9 & 67.2\\
R+E cam & 89.9 & 77.1 & 80.1 & 67.0\\
R+E pose+cam & 90.4 & 77.9 & 82.2 & 68.9 \\
\textbf{R+E (RR+) pose+cam} & \textbf{93.1} & \textbf{89.3} & \textbf{85.2} & \textbf{74.8} \\
\hline
\end{tabular}
\caption{Comparison results in Market-1501 and Duke-reID. Our method with controlled pose and camera bias \textit{R+E pose+cam}  outperforms all GANs and adversarial learning  methods. Yet, when combined with re-ranking  \textit{R+E(RR+) pose+cam} the performance further increases. }
\label{tab:compMark}
\end{table}

\begin{table}[h]
\begin{tabular}{lllll}
 & \multicolumn{2}{c}{CUHK03 (L)} & \multicolumn{2}{c}{CUHK03 (D)} \\
Method & Rank1 & \multicolumn{1}{l|}{mAP} & Rank1 & mAP \\ \hline
LOMO+XQDA~\cite{liao2015person} & 14.8 & \multicolumn{1}{l|}{13.6} & 12.8& 11.5\\ 
SVDNet~\cite{sun2017svdnet} & - & \multicolumn{1}{l|}{-} & 41.5& 37.6\\
PCB~\cite{sun2017beyond} & - & \multicolumn{1}{l|}{-} & \textbf{63.7} & \textbf{57.5} \\ 
\hline
\textit{GAN} &  & \multicolumn{1}{l|}{} &  &  \\
Pose-transfer~\cite{liu2018pose} & 45.1 & \multicolumn{1}{l|}{42} & 41.6 & 38.7 \\ 
\hline
\textit{Attention} &  & \multicolumn{1}{l|}{} &  &  \\
ACCN (D+)~\cite{xuattention} & \textbf{81.9} & \multicolumn{1}{l|}{\textbf{81.61}} & \textbf{79.1} & \textbf{78.4} \\
HA-CNN~\cite{li2018harmonious} & 44.4 & \multicolumn{1}{l|}{41} & 41.7 & 38.6 \\
Mancs~\cite{wang2018mancs}  & 69 & \multicolumn{1}{l|}{63.9} & 65.5 & 60.5 \\ 
\hline
\textit{Background bias} & - & \multicolumn{1}{l|}{-} &  &  \\
MaskG~\cite{song2018mask} & 50.1 & \multicolumn{1}{l|}{50.2} & 46.7 & 46.9 \\
\hline
\textit{Our method}& - & \multicolumn{1}{l|}{-} &  &  \\
Baseline  & 60.7 & \multicolumn{1}{l|}{72} & 58 & 69.2\\ R+E pose & 65.1 & \multicolumn{1}{l|}{75.3} & 62.3 & 72.8
\\ \textbf{R+E (RR+) pose}  & \textbf{75.5} & \multicolumn{1}{l|}{\textbf{83.9}} & \textbf{73.4} & \textbf{81.5}

\\ \hline
\end{tabular}
\caption{Comparative evaluation in CUHK03. Our framework \textit{R+E pose} and with re-ranking incorporated \textit{R+E (RR+) pose}  report competitive results on this benchmark.}
\label{tab:compCUHK03}
\end{table}
\noindent\textbf{Market-1501 and DukeMTCM-reID.}  In table \ref{tab:compMark} we observe that our proposed method that learns to reduce the pose and camera bias \textit{R+E pose+cam} improves over the existing GAN based methods, with Rank1 of $90.4\%$ on MARKET-1501 and $82.3\%$ on DUKEMTCM-reID.
Compared  to the other adversarial learning solutions, we also outperform DR-GAN~\cite{zheng2017unlabeled} by a large margin, i.e. by $3.7\%$ in MARKET-1501 and by $5.3\%$ in DUKE. Note that DR-GAN~\cite{zheng2017unlabeled} was originally designed for disentangling representations in face identification task and this results reveals the limitation of the method to generalize for articulated bodies. We also compare to FD-GAN~\cite{ge2018fd}, which addresses the pose bias. Our method shows comparable results in MARKET-1501 improving slightly the mAP from $77.7\%$ to $77.9\%$, and giving  Rank1 score of $82.2\%$ in DukeMTCM, which is higher by $2.2\%$ than FD-GAN. 
In addition, we outperform MaskG~\cite{song2018mask} by $6.6\%$, which  also tries to address the background bias. These results show the advantage of our method which can deal with different types of bias and improves upon methods targeting specific bias.
Moreover, our method  performs as well as the attention based networks, that adopt external training data and random erasing augmentation technique. 
Finally, with re-ranking incorporated \textit{R+E (RR+) pose + cam} our method with Rank1 $93.1\%$ in Market-1501 and $85.2\%$ in Duke, outperforms most of the methods in this comparative evaluation.

\noindent\textbf{CUHK03} with new $700/767$ protocol is used to obtain results in table \ref{tab:compCUHK03} for both, labeled (L) and detected (D) splits. Our framework (\textit{R+E pose}) shows very competitive scores and outperforms MaskG~\cite{song2018mask}  and Pose-transfer~\cite{liu2018pose}, which address background and pose bias, respectively. Incorporating re-ranking \textit{R+E pose (RR+)} leads to Rank1 of $75.5\%$ in labeled and $73.4\%$ in detected splits, outperforming by a large margin all the compared approaches, except $ACCN (D+)$, which uses external data for training. 
\newline
\noindent\textbf{CropCUHK03}. Finally, table \ref{tab:compcrop} shows results for our method in partial re-ID problem on CropCUHK03 and compares it against other recent techniques.  
In particular, Partial Match Net~\cite{iodice2018partial} was specifically designed to handle partial views by additional networks dealing with alignment and hallucination of missing parts.
Our method  (R+E body part + pose) significantly outperforms  Partial Match Net in Rank1  from $31.3\%$ to $43.5\%$, improving the state of art result on this dataset.  The results are reported for the same evaluation settings. Moreover, incorporating re-ranking further boosts Rank1 performance to  $54.3\%$. \newline

\begin{table}[t]
\centering
\begin{tabular}{lll}
       & \multicolumn{2}{l}{\footnotesize{CropCUHK03}} \\
                 Method            & R1                & mAP                        \\ \hline 
Res50           & 9.6               & 9.3                             \\
Res50 (RR+)    & 11.6              & 12.8                        \\
Res50+XQDA            & 16.6              & 14.5                   \\
Res50+XQDA (RR+)    & 18.6              & 20                          \\
PCB bas.         & 25.9              & 16.5                          \\
\textbf{PartialMatchNet}~\cite{iodice2018partial}    & \textbf{31.3}              & \textbf{21.2}                 \\ \hline
Baseline~\cite{hermans2017defense}             & 39.6             & 52.6                      \\
R+E body part  & 42.4             & 55.2                      \\
R+E pose                  & 42.1              & 54.7                      \\
R+E body part+pose                 & 43.5              & 56.3                         \\
\textbf{R+E body (RR+) part+pose}              & \textbf{54.3}& \textbf{65.6}                          \\
\hline
\end{tabular}
\caption{Comparative evaluation in CropCUHK03. Our approach outperforms  the other methods evaluated on this dataset. }
\label{tab:compcrop}
\end{table}

\section{Conclusion}
We proposed an approach to person re-ID based on adversarial training to extract person representation focused on ID specific cues while allowing to control the influence of bias that many re-ID benchmarks suffer from. Via bias-reducing and bias-enhancing branches we can optimize the balance of features related to re-ID and to the bias. The main novelty is in the training strategy that can be easily incorporated in other recognition problems.
We have analyzed the impact of various type of bias in re-ID task, i.e., pose, body part, camera view presenting a methodology applicable to different bias factors.
Furthermore, we demonstrate that leveraging two opposite training processes (bias-reducing and enhancing) which either suppress or emphasize bias related features, gives more robust ID representations due to the complementary content of the features.
We carried out an extensive evaluation on four different benchmarks in comparison to a number of recently proposed methods. The proposed approach consistently improves the re-ID performance compared to other state of the art techniques.   

\end{document}